%% file: main.tex
\documentclass{article}

\usepackage[final]{ap_2021}

\usepackage[utf8]{inputenc} 
\usepackage[T1]{fontenc}    
\usepackage{hyperref}       
\usepackage{url}            
\usepackage{booktabs}       
\usepackage{amsfonts}       
\usepackage{nicefrac}       
\usepackage{microtype}      

\usepackage{amsmath}
\usepackage{array}
\usepackage{mathabx}
\usepackage{csquotes}

\usepackage{graphicx}
\usepackage{arydshln}
\usepackage{multirow}
\usepackage{caption}
\usepackage{subcaption}

\title{An Approach to Inference-Driven \\
       Dialogue Management within a Social Chatbot}
\author{Sarah E. Finch\footnotemark[1], James D. Finch\footnotemark[1], Daniil Huryn, William (Mack) Hutsell, \\
\textbf{Xiaoyuan (Sophy) Huang, Han He, Jinho D. Choi\footnotemark[2]}\\
Department of Computer Science, Emory University \\
Atlanta, GA 30324 USA \\
\texttt{\{sarah.fillwock, james.darrell.finch, daniil.v.huryn, mack.hutsell,\}} \\
\texttt{\{sophy.huang, han.he, jinho.choi\}@emory.edu} \\
}

\begin{document}

\maketitle

\renewcommand{\thefootnote}{\fnsymbol{footnote}}
\footnotetext[1]{Team leads.}
\footnotetext[2]{Faculty advisor.}
\renewcommand{\thefootnote}{\arabic{footnote}}

\begin{abstract}
\input{sections/abstract}
\end{abstract}

\input{sections/introduction}
\input{sections/related_work}

\input{sections/semantic_representation}
\input{sections/graph_representation}

\input{sections/nlu}
\input{sections/dialogue_management}
\input{sections/nlg}
\input{sections/interaction_design}
\input{sections/discussion}
\input{sections/conclusion}

\bibliography{refs}
\bibliographystyle{acl}

\end{document}

%% file: sections/abstract.tex
We present a chatbot implementing a novel dialogue management approach based on logical inference.
Instead of framing conversation a sequence of response generation tasks, we model conversation as a collaborative inference process in which speakers share information to synthesize new knowledge in real time. Our chatbot pipeline accomplishes this modelling in three broad stages. The first stage translates user utterances into a symbolic predicate representation. The second stage then uses this structured representation in conjunction with a larger knowledge base to synthesize new predicates using efficient graph matching. In the third and final stage, our bot selects a small subset of predicates and translates them into an English response. This approach lends itself to understanding latent semantics of user inputs, flexible initiative taking, and responses that are novel and coherent with the dialogue context.

%% file: sections/introduction.tex
\section{Introduction}

Human-computer chat is one of the most difficult challenges in modern AI research. The goal is for a chatbot be able to engage a user in long-form conversation on a variety of topics in a way that is satisfying to the user. User satisfaction is affected by a range of factors, including the interestingness of bot responses and topic of conversation chosen by the bot. However, we believe the outstanding challenge of achieving satisfying conversations is the ability of chatbots to meet basic expectations of the user and follow the unwritten rules of human conversation.

Instead of framing conversation a sequence of response generation tasks, we view conversation as a collaborative inference process in which each conversation participant---human or chatbot---has a partially disjoint knowledge base and a goal of expanding that knowledge. Each utterance in the dialogue is a transfer of information from one converser to another across a noisy speech channel. In addition to communicating unique prior knowledge, a key feature of conversation is the ability of each converser to synthesize new information from discussed topics in real-time. This leads to interactions where the participants are able to collaborate to generate novel information that coheres with their prior knowledge, but was not known to either participant before the conversation began.

Following this framing of the problem, we present a novel dialogue framework and implemented chatbot that dynamically infers new knowledge during conversation by synthesizing prior knowledge with information presented by the user. Our dialogue management strategy uses this inference process both to understand latent meanings of the user's utterances, and to create interesting and relevant responses that logically follow from the dialogue history. Although this kind of inference-driven dialogue management could be realized in a variety of ways, we choose a symbolic state representation and inference algorithm in order to maximize the controllability of system behaviors and transparency of its reasoning process. 

\begin{figure*}[!htbp]
\centering
\includegraphics[width=\textwidth]{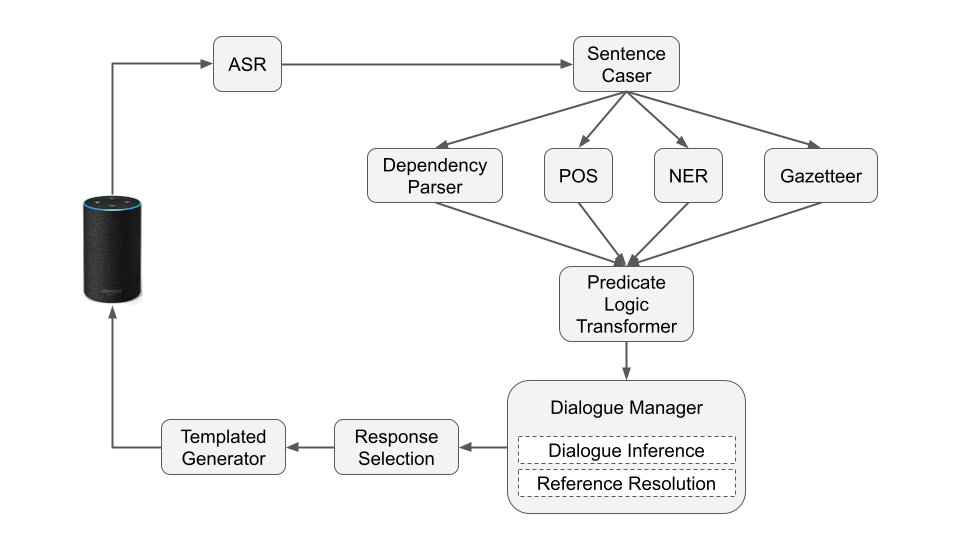}
\caption{Overall system architecture from reception of user utterance to production of system response.}
\label{fig:system_arch}
\end{figure*}

Broadly, our chatbot's pipeline (shown in Figure \ref{fig:system_arch}) is decomposed into three stages. First, a natural language understanding stage (NLU) where natural language utterances from a human are translated into a predicate logic representation. The second stage is the inference process that generates new predicates that are entailed from a combination of prior knowledge and information consumed via NLU. Third, a small subset of predicates are selected and translated into natural language as a response to the user. This report describes our details of our approach and implemented pipeline.

%% file: sections/related_work.tex
\section{Related Work}

Current approaches to chit-chat-oriented dialogue systems focus on neural language modelling architectures, with the goal being that the dialogue system will learn to mimic human agents by training on human-human conversations \citep{adiwardana2020, roller2020}. Although neural approaches have been successful when applied to many tasks, they have significant limitations when applied to the dialogue setting. These neural language modelling approaches to dialogue management often hallucinate incorrect information or seem to forget information that is included in the conversation context. This is detrimental for longer, chit-chat oriented conversations because it leads to frequent misunderstandings and frustration on the user's part when presented with an inconsistent and contradictory dialogue partner \citep{roller2020survey}. In addition, neural approaches afford little control in terms of conversation content, flow or style, which makes it challenging to correct issues or guarantee certain dialogue behaviors will be performed, which is unacceptable in a production setting.

In light of this, it is unsuprising that recent iterations of the Alexa Prize competition have presented socialbots that rely on rule-based dialogue management, through architectures such as state-machines \citep{gabriel2020}. These rule-based dialogue approaches allow less broad understanding than language modelling approaches but afford far more controllability in terms of the conversation experience and if designed properly, do not have problems of forgetting or hallucinating information. However, rule-based dialogue management systems such as state-machine approaches require the developer to handcraft a rule for each possible state of the conversation that is desired to be handled. This results in dialogue managers that are overly rigid in terms of the dialogue flows that they support, since a developer would have had to anticipate each possible conversation path and generate the appropriate supporting content during development.

Inspired by the general language understanding accomplished by neural models and the controllability of state-machines, our dialogue system aims to incorporate the best of both paradigms by managing dialogue in a consistent yet flexible way using an explicit dialogue state representation composed of fine-grained predicate logic. We capitalize on the general understanding of neural NLP models by deriving predicate logic representations of user utterances, but avoid the forgetfulness and hallucination problems of neural dialogue approaches due to our explicit state representation. Although it is symbolic, our state representation operates on a fine-grained predicate level rather than representing the context as a single state like in a state-machine, which allows our dialogue management to be far more flexible than other rule-based approaches. The next response is simply a fluid continuation of all of the recently introduced concepts. In addition, our dialogue management is much more powerful than other rule-based approaches with the incorporation of our symbolic inference engine that enables easy encoding of reasoning, including common-sense and more complex reasoning capabilities

%% file: sections/semantic_representation.tex
\section{Semantic Representation}
\label{semantic_representation}

Our approach relies on an explicit predicate logic representation to encode both the semantic meaning of natural language and information in the chatbot's knowledge base.
This semantic representation is composed of a set of symbols called \textit{concepts}.
Each concept represents a distinct item, attribute, event, or idea in the world.
Additionally, we represent relationships between concepts using predicate-argument structures, similar to how they are used in other semantic representation systems such as First Order Logic or Abstract Meaning Representation. We call a collection of concepts and their predicate structures a \textit{Concept Graph} (CG), and these CG's are used by many stages of our pipeline to represent our chatbot's knowledge and intermediate processing.

\subsection{Concept Graph}

Formally, we define a Concept Graph $K$ as a 4-tuple $K = (C_K, P_K, \pi_K, \Omega_K)$.

$C_K$ is a finite set of symbols representing concepts. 

$P_K$ is a subset of concepts representing predicates, $P_K \subset C_K$.

$\pi_K$ is a function $\pi_K: P_K \rightarrow C_K \bigtimes (C_K \bigcup \{\epsilon\})$
representing the argument structure of predicates in $P_K$ ($\epsilon$ is a special "null" symbol used for representing unary predicates).

$\Omega_K$ is a directed acyclic graph $\Omega_K = (C_K, E_K)$ representing the type ontology, with all concepts $C_K$ as nodes and edges $E_K$ representing some acyclic type hierarchy.

Additionally, we define two constructs, $T_K$ and $\tau_K$, derivable from the ontology definition $\Omega_K$.

$T_K$ is the set of all types, $T_K = \{t \in C_K : t$ has at least one in-edge in $\Omega\}$

$\tau_K$ is a function defining all types of a concept, $\tau_K(c) = \{t \in T_K : t $ is some ancestor of $c$ in $\Omega\}$

As an example usage of our CG, suppose we want to represent the concepts introduced by the English sentence "Tom watched a dog". 
We will define a CG called $X$ to represent the semantic meaning of this sentence in our example. According to the definition above, $X = (C_X, P_X, \pi_X, \Omega_X)$. 
Since the sentence presents three main concepts: Tom, the dog, and the watch action done by Tom, each of these concepts is represented as a unique symbol in the concept set. If Tom, the dog, and the watch action are represented by the symbols $tom$, $d$, and $w$ respectively, then $\{tom, d, w\} \subseteq C_X$. 

Capturing the relationships between these concepts as conveyed in the sentence requires composing them into a predicate-argument structure. In our example, $tom$ and $d$ are related by a watching action $w$.
We can use $w$ as a concept representing a relationship between $tom$ and $d$ by defining it as a predicate, $P_X = \{w\}$.
Then, we define $\pi_X$ such that $\pi_X(w) = (tom, d)$ to encode the idea that $tom$ is the first argument of predicate $w$ and $d$ is the second argument.

Finally, a major source of meaning in concepts mentioned by natural language is background information known about the types (also known as schemas) of each concept.
In the example sentence, we know that $tom$ is a person, $d$ is a dog, and $w$ is a watch event. Knowledge and inferences that apply in general to people should apply to $tom$, those applying to dogs should apply to $d$, and so on. We can encode abstract type concepts as any other concept in $C_K$. If $person$, $dog$, and $watch$ represent three type concepts, $\{person, dog, watch\} \subset C_K$. This gives us a relatively complete set of concepts to represent the sentence, $C_K = \{tom, d, w, person, dog, watch\}$. Encoding the three new concepts as type information is done by defining $\Omega$ such that $E_K = \{(tom, person), (d, dog), (w, watch)\}$. In this way, the representation captures the idea that instance concepts such as $d$ are members of an abstract collection of similar objects, $\tau_X(d) = \{dog\}$.

By encoding the conversation context and the bot's knowledge with a direct predicate form, we maintain direct explainability and controllability over all of the bot's decision-making processes. It is easy to add new knowledge into the system, when desired by a developer. To this end, we created a custom compiler that translates structured text into CG's. For ease of explanation, we will sometimes use the following syntax defined in our compiler to present predicates and CG information within this report, since the notation is more concise than the expanded formal definitions.

In general, the form:

\begin{equation}
    predicate/predicate\_type(arg0, arg1)
\end{equation}

represents four concepts, $\{predicate, predicate\_type, arg0, arg1\} \subseteq C$ with type information $\{predicate\_type\} \subseteq \tau(predicate)$, and argument structure $\pi(predicate) = (arg0, arg1)$.

Similarly, the form:

\begin{equation}
    entity/entity\_type()
\end{equation}

represents concepts $\{entity, entity\_type\} \subseteq C$, with type information $\{entity\_type\} \subseteq \tau(entity)$.

Table \ref{tab:semantic_translations} provides additional examples for typical translations of English sentences into our semantic representation used in the system.

\begin{table*}[htbp!]
\setlength\extrarowheight{4pt}
\centering\resizebox{\textwidth}{!}{
  \begin{tabular}{l||l}
  \bf English Sentence & \bf Predicate Logic\\
  \hline
  \hline
  \multirow{2}{*}{I ran on the treadmill.} & \tt r/run(user) time(r, past) \\ 
  & \tt on(r, t) type(t, treadmill) \\
  
  \hline
  \multirow{2}{*}{I like watching action movies.} & \tt l/like(user, watch(user, m)) time(l, now) \\
  & \tt type(m, movie) type(m, group) action(m) \\
  
  \hline
  Your dog is sweet. & \tt s/sweet(d) time(s, now) type(d, dog) possess(bot, d)\\
  
  \hline
  I am a math teacher. & \tt b/be(user, t) time(b, now) type(t, teacher) of(t, math) \\
  
  \hline
  \multirow{3}{*}{My grades fell quickly after I stopped studying.} & \tt f/fall(g) time(f, past) type(g, grade) \\
  & \tt type(g, group) possess(user, g) quick(f) \\
  & \tt after(f, s/stop(user, study(user)) time(s, past) \\
  
  \hline
  I didn't eat lunch yet. & \tt e/eat(user, l) type(l, lunch) not(e) time(e, past)\\
  
  \hline
  I should eat lunch. & \tt e/eat(user, l) type(l, lunch) should(e) time(e, now)\\
  
  \hline
  \multirow{2}{*}{What musical instrument do you play?} & \tt p/play(bot, i) type(i, musical\_instrument) time(p, now) \\
  & \tt request(user, i) \\
  
  \hline
  \multirow{3}{*}{Did you like the book I gave you?} & \tt l/like(bot, b) type(b, book) time(l, past) \\
  & \tt g/give(user, b) recipient(g, bot) time(g, past) \\
  & \tt request\_truth(user, l)\\

  \end{tabular}}
  \caption{Example translations between English sentences and their corresponding predicate logic.}
  \label{tab:semantic_translations}
\end{table*}

\subsection{Inference rules}
\label{inference_rules}

Similar to predicate logic, our semantic representation has the notion of an implication, which can be used to assert new predicates from an existing set of facts. 

Inference rules follow the format $precondition \rightarrow postcondition$.
Each precondition is defined as a tuple $(Q, V)$ where $Q$ is a CG and $V \subseteq C_Q$ is a set of variables. Preconditions are \textit{satisfied} by a CG $K$ by finding a replacement of each variable $x_i \in X$ with some concept $c \in C_{K}$ such that $C_{Q} \subseteq C_{K}$, \hspace{2mm} $\forall p \in P_Q : \pi_K(p) = \pi_Q(p)$, and $\forall c \in C_Q : \tau_Q(c) \subseteq \tau_K(c)$. Each assignment of variables to concepts from $C_K$ is considered a solution to the inference rule. Since inference rules represent logical implications, each solution can be used to replace the appropriate variables in the CG postcondition of the satisfied rule, thus generating a new set of logically consistent concepts.

As an example, consider the following inference rule:

\begin{equation}
    W/wag(X/dog(), Y/tail()) \quad \rightarrow \quad happy(X)
\end{equation}

and the following knowledge:

\begin{equation}
\begin{split}
    w1/wag(fido/dog(), t1/tail()) \\
    w2/wag(spot/dog(), t2/tail()) \\
    w3/wag(dash/cat(), t3/tail())
\end{split}
\end{equation}

There are two valid solutions in the knowledge set that satisfy the inference rule. One solution is found from the unification $\{(W, w1), (X, fido), (Y, t1)\}$. The other is found from the unification $\{(W, w2), (X, spot), (Y, t2)\}$. There is no solution involving $dash$ since $dash$ cannot unify with $X$ while also satisfying the type requirement $dog \in \tau_Q(X)$.

As a result of the two valid solutions, two new predicates can be inferred by applying the rule to the knowledge. By filling the variables in the postconditions of rule with the identified variable assignments, the new knowledge $happy(fido)$ and $happy(spot)$ are produced.

%% file: sections/graph_representation.tex
\section{Graph Representation and Matching}
\label{graph_matching}

The semantic representation described in section \ref{semantic_representation} is a flexible and powerful way to reason with information by applying inference rules to CG's. However, in practice, quickly accessing information in a CG representation and computing unifications of inference rule variables can be a computationally demanding challenge. To address this, our implementation uses efficient graph data structures and matching algorithms to represent and manipulate CG's.

A CG $K$ in our semantic representation can be easily represented in practice as a directed labelled graph $G = (V, E)$. Each concept is represented as a node in the graph $V = C_K$. Directed edges then represent both the argument attachment structures of the predicates defined by $P_K$ and $\pi_K$, as well as the type hierarchy $\Omega_K$. To represent argument attachments, for each $p \in P_K$ defining $p/t(s, o)$, $E$ contains corresponding labelled edges $(p,s,ARG0)$, $(p,o,ARG1)$. Similarly, each concept $c$ with types $\tau_K(c)$, each $t \in \tau_K(c)$ forms an edge $(c,t,T)$.

\begin{figure*}[!htbp]
\centering
\includegraphics[width=0.75\textwidth]{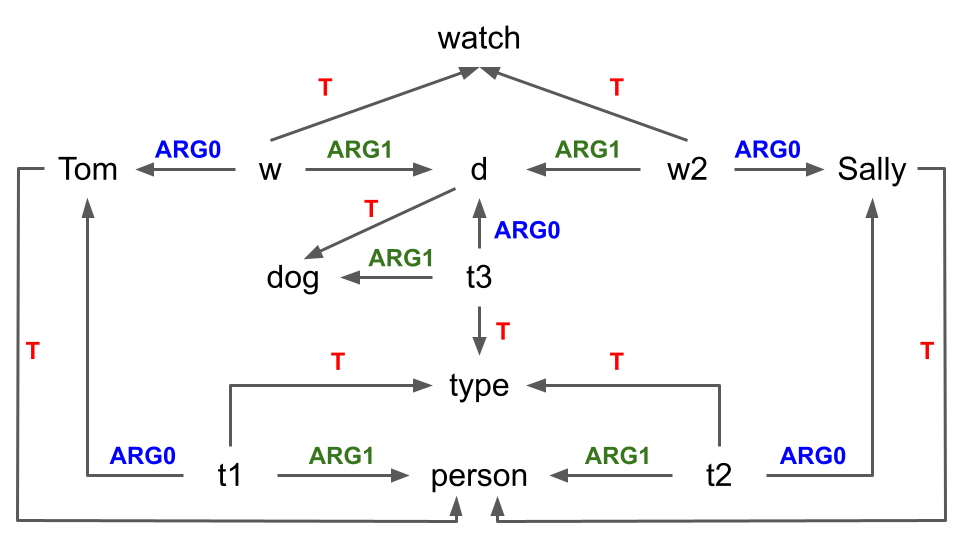}
\caption{Directed labelled graph representations for the actions of (w) Tom watching a dog and (w2) Sally watching the same dog.}
\label{fig:graph_ex}
\end{figure*}

Figure \ref{fig:graph_ex} illustrates the directed labelled graph that represents the following CG:

\begin{align*}
    &w/watch(Tom, d) \\
    &t1/type(Tom, person) \\
    &w2/watch(Sally, d) \\
    &t2/type(Sally ,person) \\
    &t3/type(d, dog)
\end{align*}

Each predicate concept \textit{w, w2, t1, t2, and t3} participates in three labelled edges in the directed graph. The subjects \textit{Tom, Sally, and d} of the type predicates \textit{t1, t2, and t3}, respectively, also have a direct T edge to the object of their type predicate.

Given this equivalence in representation between our predicate logic and directed graphs, we treat the task of finding solutions to inference rule preconditions as a graph pattern matching problem. The goal of this matching problem is to find a set of solutions to rule preconditions from a set of concepts in a knowledge base. The precondition of each inference rule $i$ is represented as a query graph $Q_i$ in the graph matching problem with a set of variable nodes $X$. The knowledge base is similarly transformed into the data graph $D$ of the graph matching problem. Since finding a subgraph of $D$ isomorphic to some $Q_i$ entails satisfying all predicate-argument attachments of the original inference rule $i$'s precondition, all data graph nodes matched to variables in the query graph compose an assignment for a valid inference rule solution.

A key challenge of using a graph matching approach to satisfy inference rules within a dialogue context is the relatively high time complexity of the graph matching problem. Our approach relies on having hundreds or thousands of inference rules which must be checked against our bot's knowledge of the current conversation in well under a second. To overcome this, we use an algorithm similar to \cite{bhattarai2019ceci} in order to match a large collection of query graphs against a data graph in parallel on a GPU architecture.

%% file: sections/nlu.tex
\section{Natural Language Understanding}
\label{nlu}

The goal of our natural language understanding system is to translate English utterances into our predicate logic representation. To do this, our NLU system follows two main steps. First, the utterance is processed by a suite of NLP models. Then, the structured outputs of our NLP models are used to construct the final predicate logic representation of the utterance as described in Section \ref{semantic_representation}. 

\input{sections/nlp_models}

\subsection{Generating Predicate Logic}
\label{gen_pred_log}

The sentence caser preprocesses the user utterance to produce a cased version, before the remaining four NLP models are then run on the cased user utterance. We then use their results to transform the user utterance into our semantic representation. 

First, we identify a set of concepts from the outputs of the gazetteer matcher, NER model, and POS model. The gazetteer matcher produces concepts using the knowledge base as described in Section \ref{gazetteer}. For each tagged span identified by the NER or POS models, we instantiate a new concept where the type of the concept is defined by the outputted tag. Since each model may identify different concepts for the same natural language span, we use a priority system to select a subset of all identified concepts. The gazetter matcher is given the highest priority such that all concepts identified by the gazetteer matcher are selected. The middle tier consists of the NER model; only those concepts identified by the NER with spans which do not overlap with any span identified by the gazetteer matcher are selected. Finally, the POS model is lowest priority and identifies concepts for any tokens not otherwise accounted for by the other two models. At the conclusion of this step, we have obtained a mapping from token spans in the user utterance to their corresponding concepts, where each token is accounted for exactly once.

As an example, let's take the following user utterance:

\begin{center}
    \begin{tabular}{l}
    $(U1)$ Tom watched the dog by the bus stop near Central Park. 
    \end{tabular}
\end{center}
 
Let's suppose that the concept outputs from each of the three models are those shown in Table \ref{concept_outputs}.

\begin{table}[htbp]
\subcaptionbox{Concepts from Gazetteer}{
\small \begin{tabular}[t]{c|l} 
    Span & Concept\\ 
    \hline
    Tom & $Tom$\\ 
    watched & $w*/watch(-, -)$\\ 
    dog & $type(d*, dog)$ \\    
    by & $b*/by(-, -)$\\
    bus stop & $type(bs*, bus\_stop$) \\ 
    near & $n*/near(-, -)$ \\ 
    &\\
    &\\
    &\\
\end{tabular}
}
\hfill
\subcaptionbox{Concepts based on NER}{
\small \begin{tabular}[t]{c|l}
    Span & Concept\\ 
    \hline
    Tom & $type(t*, per)$ \\ 
    Central Park & $type(cp*, loc)$ \\
    &\\
    &\\
    &\\
    &\\
    &\\
    &\\
    &\\
\end{tabular}
}
\hfill
\subcaptionbox{Concepts based on POS}{
\small \begin{tabular}[t]{c|l}
    Span & Concept\\ 
    \hline
    Tom & $type(t*, nnp)$\\ 
    watched & $type(w*, vbd)$\\ 
    dog & $type(d*, nn)$\\     
    by & $type(b*, in)$\\ 
    bus & $type(b*, nn)$\\ 
    stop & $type(s*, nn)$\\ 
    near & $type(n*, in)$\\ 
    Central & $type(c*, nnp)$\\ 
    Park & $type(p*, nnp)$ \\ 
\end{tabular}
}
\hfill
\caption{Concept outputs for user utterance $U1$}
\label{concept_outputs}
\end{table}

The asterisk $*$ denotes a new entity or predicate instance produced as the focal concept output for that span, where new entity instances have some supporting type predicate instance and new predicate instances have some supporting predicate-argument structure with $-$ denoting the expected but currently unfilled argument slots.

From just the top-priority gazetteer, all tokens are accounted for except for "Central" and "Park". For the missing tokens, the NER model is considered next, based on the priority system. Since the NER model identified a concept for the span "Central Park", all concepts have now been identified for all tokens in the user utterance, with the final span-to-concept mapping shown in Table \ref{tab:final_map}.

\begin{table}[htbp]
    \centering
    \begin{tabular}{c|l}
        Span & Concept\\ 
        \hline
        Tom & $Tom$\\ 
        watched & $w*/watch(null,null)$\\ 
        dog & $type(d*, dog)$ \\    
        by & $b*/by(null, null)$\\
        bus stop & $type(bs*, bus\_stop$) \\ 
        near & $n*/near(null,null)$ \\ 
        Central Park & $type(cp*, loc)$ \\
    \end{tabular}
    \vspace{3mm}
    \caption{Final span-to-concept mapping based on all 3 NLP models adjusted by the priority system}
    \label{tab:final_map}
\end{table}

Once the mapping of spans to concepts has been identified, the next task is to connect the concepts together with appropriate predicate-argument structures. We use the outputs of the POS tagger and the dependency parser to inform the predicate-argument structure creation through the application of transformation rules. 

Transformation rules are each of the form $parse\_pattern \rightarrow A$. $parse\_pattern$ is a directed labelled graph where a subset of its nodes are variables, intended to match subgraphs of dependency parses. $A$ is a set of logical attachments represented as 3-tuples $(span_a, slot, span_b)$, where $span_a$ and $span_b$ are spans corresponding to variables in the precondition, and $slot \in \{T, ARG0, ARG1\}$ corresponding to an attachment label in the graphical form of our semantic representation. Based on the solutions found by matching each transformation rule's parse pattern against the dependency parse graph outputted by our model, a set of span-based attachments are identified. These span-based attachments are easily converted to attachments between concepts by replacing the spans with their corresponding concepts according to the mapping from the previous step.

\begin{figure*}[!htbp]
\centering
\includegraphics[width=\textwidth]{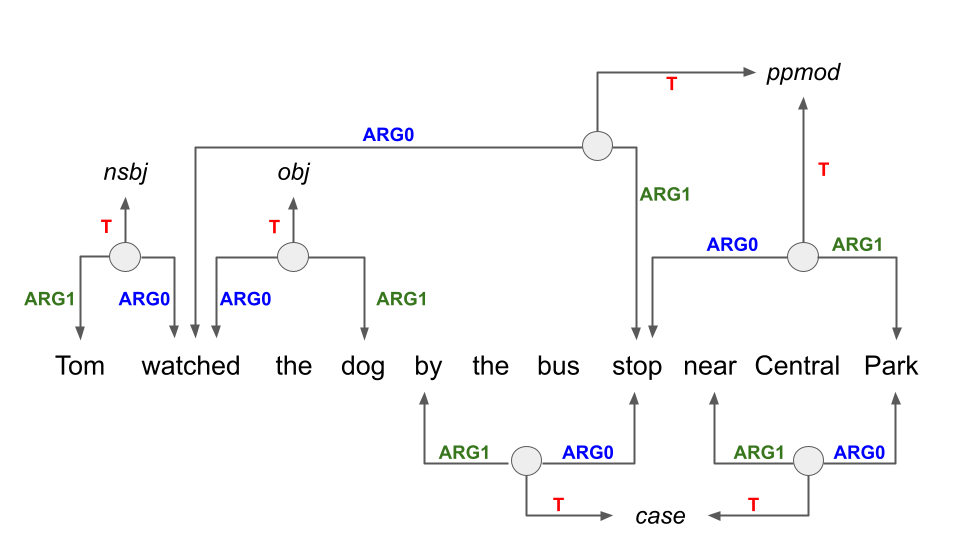}
\caption{Directed labelled graph representation of the dependency parse outputs for $U1$ which are relevant to the example.}
\label{fig:parse_graph_ex}
\end{figure*}

We use the graph matching setup from Section \ref{graph_matching} for the application of transformation rules. The dependency parse is represented as a directed labelled graph where the nodes are token spans and dependency link labels, and the edges denote the head-child relationships of identified dependency structures. The outputs from the POS tagger augment the dependency parse graph by adding an additional edge from each token span to its POS tag.

Note that the attachment identification based on the dependency parser may produce span attachments that do not directly correspond to a span in the span-to-concept mapping, due to the utilization of concept identification models that are able to recognize spans of length greater than one. To account for this, spans outputted in the attachment tuples are matched to the span that contains them in the span-to-concept mapping, which may be exactly themselves or some multi-token span of which they are a subset.

Continuing with the example utterance $U1$, its parse graph can be found in Figure \ref{fig:parse_graph_ex}. For readability, only those dependency parse outputs that are relevant to the example have been included and all type edges denoting POS tags have been excluded.

\begin{figure}[htp]
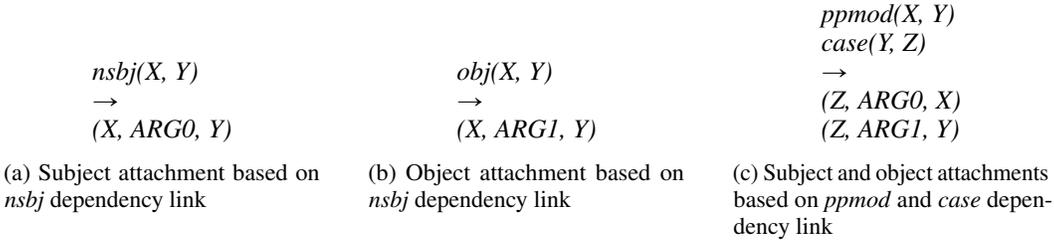

\subcaptionbox{Subject attachment based on \textit{nsbj} dependency link}[0.3\textwidth]{
\begin{tabular}[t]{l} 
    \textit{nsbj(X, Y)} \\
    $\rightarrow$ \\
    \textit{(X, ARG0, Y)}
\end{tabular}
}
\hfill
\subcaptionbox{Object attachment based on \textit{nsbj} dependency link}[0.3\textwidth]{
\begin{tabular}[t]{l} 
    \textit{obj(X, Y)} \\
    $\rightarrow$ \\
    \textit{(X, ARG1, Y)}
\end{tabular}
}
\hfill
\subcaptionbox{Subject and object attachments based on \textit{ppmod} and \textit{case} dependency link}[0.3\textwidth]{
\begin{tabular}[t]{l} 
    \textit{ppmod(X, Y)} \\
    \textit{case(Y, Z)} \\
    $\rightarrow$ \\
    \textit{(Z, ARG0, X)} \\
    \textit{(Z, ARG1, Y)}
\end{tabular}
}
\hfill
\caption{Transformation rules applicable to user utterance $U1$}
\label{transformation_rules}
\end{figure}

Let's take the set of transformation rules in Figure \ref{transformation_rules}.

Rules (a) and (b) specify the subject and object attachments, respectively, for predicate instances that are derived from subject-verb-object grammatical forms. Rule (c) dictates both the subject and object attachments for a predicate instance which represents a prepositional clause.

There are four solutions (and thus four span-attachment sets) found from the application of these transformations rules to the parse graph of $U1$: one from rule (a), one from rule (b), and two from rule (c), as shown in Table \ref{tab:transformation_results}.

\begin{table}[htp]
\centering
    \begin{tabular}[t]{c|l|l|l} 
        Rule & Solution & Span Attachments & Concept Attachments\\
        \hline\hline
        \multirow{2}{*}{(a)} & X: "watched" & \multirow{2}{*}{("watched", ARG0, "Tom")} & \multirow{2}{*}{($w$, ARG0, $Tom$)} \\
        & Y: "Tom" & & \\
        \hline
        \multirow{2}{*}{(b)} & X: "watched" & \multirow{2}{*}{("watched", ARG1, "dog")} & \multirow{2}{*}{($w$, ARG1, $d$)} \\
        & Y: "dog" & & \\
        \hline
        \multirow{3}{*}{(c)} & X: "watched" & ("by", ARG0, "watched") & ($b$, ARG0, $w$) \\
        & Y: "stop" & ("by", ARG1, "stop") & ($b$, ARG1, $bs$) \\
        & Z: "by" & & \\
        \hline
        \multirow{3}{*}{(c)} & X: "stop" & ("near", ARG0, "stop") & ($n$, ARG0, $bs$) \\
        & Y: "park" & ("near", ARG1, "park") & ($n$, ARG1, $cp$) \\
        & Z: "near" & & \\
    \end{tabular}
\vspace{3mm}
\caption{Transformation rule results}
\label{tab:transformation_results}
\end{table}

Given the span-to-concept mapping from earlier as well as these identified span-attachment sets, the final concept-attachment sets can also be derived. All spans are directly in the span-to-concept mapping, with the exception of "stop" and "Park". Since "stop" and "Park" were each recognized as part of a multi-token span, "bus stop" and "Central Park" respectively, their corresponding span in the span-to-concept mapping is the multi-token span that they are a part of. After translating the span-attachments into the appropriate concept-attachments based on the previous considerations and then performing the indicated attachments, the final predicate logic form of the user utterance becomes:

\begin{align*}
    &type(d, dog) \\
    &type(bs, bus\_stop) \\
    &type(cp, loc) \\
    &w/watch(Tom, d) \\
    &b/by(w, bs) \\
    &n/near(bs, cp) \\
\end{align*}

\subsection{References}
\label{references}

One of the most difficult challenges in dialogue is the idea of partial information, where a concept is being referred to that either does not have an explicit identifier or the explicit identifier is unknown by one conversational partner. Common cases are pronouns like \textit{it} or \textit{he}, referential language like \textit{the dog in the park}, and questions like \textit{What is your name}. In these cases, it is assumed that both parties are aware of the concept being referred to; however, it is up to the listener to disambiguate the exact concept that is being referred to, based on the information and properties shared by the speaker that discriminate candidate instances from one another. To overcome this challenge, our semantic representation has a reference representation, where a reference concept is defined as a variable that participates in a collection of predicates and properties that describe it. The reference variable is resolved by identifying the concept that satisfies all of the references constraints once it is unified with the reference. 

References are defined in our system as a tuple of the form $(focus, variables, constraints)$ where $constraints$ is a CG, $variables \subseteq C_{constraints}$, and $focus \in variables$ represents the intended referent. In this paradigm, a reference specification defines a set of referent candidates as the set of all concepts which are assigned to the $focus$ in some satisfaction of the $constraints$. Computing reference satisfactions can be accomplished using the approach describe in Section \ref{graph_matching} by treating the $constraints$ as a query graph.

Unlike inference rules, references can be generated in runtime via both the bot's reasoning process or as some translation of the user utterance into our predicate representation. To facilitate this, we use two special predicate types $REF$ and $VAR$ which links a reference focus with its constraints and variables, respectively. In the case of interpreting a reference from the user utterance, the parse transformation rules (described in Section \ref{gen_pred_log}) can optionally encode a set of REF and VAR predicates in their postconditions to set up a reference representation in addition to an attachment set $A$. Similarly, inference rules used in the dialogue inference step can define $REF$ and $VAR$ predicates in their postconditions in the same manner as specifying any other predicates.

During the reference resolution step of our pipeline, all $REF$ and $VAR$ predicates in working memory are collected in order to run graph matching for the references that they define. This produces a set of candidate reference resolution pairs $(r_i, r_j)$ where each $r_i$ is the focus concept of some reference and its corresponding $r_j$ is a candidate referent for that reference. We then merge together each $r_i$ with the $r_j$ that has the greatest relevance to the current conversational context (relevance is represented by the \textit{salience} feature described in Section \ref{working_memory}). 

%% file: sections/nlp_models.tex
\subsection{NLP Models}

Five NLP models are utilized in the natural language understanding system. Although all of the models operate on the same input utterance, they each produce a unique set of information that is utilized later in the pipeline for further processing. 

Sections \ref{pos}, \ref{ner}, and \ref{dep} describe models provided through the Emory Language and Information Toolkit (ELIT)\footnote{\url{https://github.com/emorynlp/elit}}. Compared to existing widely used toolkits, ELIT features an efficient Multi-Task Learning (MTL) framework that supports the many sub-tasks it provides. Each sub-task shares the Transformer-Encoder to get the token-level embedding of the input utterance, but has an independent state-of-the-art decoder to which the token embeddings are fed in parallel.

\subsubsection{Sentence Caser}

The sentencer caser applies casing to tokens in the input utterance such that the most likely overall sentence casing is achieved according to a provided statistical distribution. We use the sentence caser from the truecaser github project\footnote{\url{https://github.com/nreimers/truecaser}}, which provides an English statistical casing distribution based on several open-source English corpora.

\subsubsection{Gazetteer Matcher}
\label{gazetteer}

The gazetteer matcher identifies all subspans of the input utterance that correspond to a concept in the knowledge base KB. It utilizes the Trie-based Aho-Corasick algorithm \citep{aho1975} provided in the python package pyahocorasick\footnote{\url{https://pyahocorasick.readthedocs.io/en/latest/}} for efficient exact string matching on any length of strings. 

The KB is constructed with a many-to-one mapping between natural language strings and concepts, such that each concept is defined with a set of strings that are used to refer to that concept. This allows for the capture of synonyms and other lexical variants of each concept. This mapping is a surjective function since each string maps to only one concept and each concept can be mapped to by more than one string. Since this approach to concept identification disallows homonyms, it only accomplishes an approximation of the true flexibility of natural language and future work involves incorporating a more sophisticated disambiguation strategy to enable better understanding.

There are three types of concepts that may be identified by the gazetteer matcher:

\paragraph{a. Entity Instances} The identified concept is a named entity instance in the KB. For example, the string "George Washington" corresponds to the concept \textit{george\_washington}, which is a person instance representing the first president of the United States. In this case, the gazetteer matcher outputs the concept directly.

\paragraph{b. Entity Types} The identified concept is an entity type in the KB. Unlike (a), this identified concept does not refer to a specific instance in the KB, but rather a type of instances. This case arises when the user mentions an unnamed entity instance; without a specific identifier that refers to the instance, the user must instead refer to the instance by its class, at minimum, to discriminate the instance to their conversational partner. For example, the string "dog" corresponds to the concept \textit{dog} which is a type in the KB and is mentioned when the user is referring to some dog relevant to the discussion. In this case, the gazetteer matcher outputs a new entity instance $e$ with its supporting structure \textit{p/type(e, c)} where $p$ is a new predicate instance with subject $e$, predicate type \textit{type}, and object $c$ which is the identified concept. 

\paragraph{c. Predicate Types} The identified concept is a predicate type in the KB. A predicate type is present in the user utterance only when the user is mentioning an instance of that predicate, since predicate types are meaningless without their argument structure specified. For instance, in the utterance "Tom watched his dog", the "watched" string corresponds to the \textit{watch} predicate type and portrays an instance of \textit{watch} where "Tom" is the subject and "his dog" is the object. In this case, the gazetteer matcher outputs a new predicate instance with the concept as its predicate type and an unfilled subject argument. If the predicate type expects an object argument according to the KB, then the new predicate instance is also given an unfilled object argument.

\subsubsection{POS Tagger} 
\label{pos}
We use a POS tagger to assign POS tags to tokens in the utterance. For efficiency, the POS decoder consists only of a linear layer, ignoring character level and case features \citep{bohnet-etal-2018-morphosyntactic,akbik-etal-2018-contextual} since they bring marginal accuracy improvements compared to their increase in latency. 

\subsubsection{NER} 
\label{ner}
An NER model is used to extract named entity spans from the user utterance. The NER decoder consists of a biaffine layer \citep{dozat:17a}. Different from \cite{yu-etal-2020-named}, we avoid using document level features and the variational BiLSTM for faster decoding speed.

\subsubsection{Dependency Parser} 
\label{dep}
A dependency parser identifies the dependency relations between tokens in the utterance. The dependency parser decoder consists of two biaffine layers \citep{dozat:17a} which are used to compute the dependent-head score matrix and label distribution. Then the Chu-Liu/Edmonds algorithm \citep{chu1965shortest, edmonds1967optimum} is applied on the score matrix to decode the results.

\subsubsection{ELIT NLP Model Performance}

\paragraph{Datasets} The ELIT models used in this competition are trained on a combination of OntoNotes 5 \citep{weischedel2013ontonotes}, BOLT English Treebanks \citep{song2019bolt, tracey2019bolt}, THYME/SHARP/MiPACQ Treebank \citep{albright2013towards}, English web treebank \citep{bies2012english}, Questionbank \citep{judge2006questionbank} and the officially released AMR 3.0 \citep{knight2020abstract}. Due to the multi-task learning setup, batches of NLP tasks are mixed together so that even if a corpus offers no annotation for some tasks, it can still be exploited by ELIT. We follow the train-dev-test splits standard to the official release of each dataset for our training and evaluation. 

\paragraph{Performance} The following metrics are used for each task - POS: accuracy, NER: span-level labeled F1, DEP: labeled attachment score. Table~\ref{tab:mtl-performance} shows the scores on the combined test sets.

\begin{table}[htbp]
\centering
\begin{tabular}{c|c|c|c}

     & POS   & NER   & DEP \\ \hline \hline
ELIT & 98.08 & 89.01 & 91.21  \\ 
\end{tabular}
\vspace{3mm}
\caption{Performance of the MTL-RoBERTa models in ELIT.}
\label{tab:mtl-performance}
\end{table}

%% file: sections/dialogue_management.tex
\section{Dialogue Management}
\label{working_memory}

\subsection{Working Memory}

A consistent challenge in human-computer dialogue is designing a computationally feasible state representation that adequately accounts for the dialogue history.
Our approach to dialogue state representation is to maintain a set of concepts that were either directly mentioned in the recent dialogue history, or are semantically related to those mentioned concepts. This approach has two major goals: 1) to prioritize the inclusion of concepts in the state representation based on their relevance to the current discussion, and 2) to manage the size of the state representation so that it is large enough to facilitate meaningful interactions, while not growing so large as to put strain on the pipeline's response latency.

Specifically, we use a CG named \textit{Working Memory} (WM) to represent the dialogue state. Working memory is initialized with a small set of predefined concepts to initialize the conversation, but is expanded by adding new concepts throughout the conversation to model relevant concepts as they are introduced. At the beginning of each turn, the CG produced by our NLU subpipeline (section \ref{nlu}) is added to WM using union operations. WM is then further expanded using a knowledge retrieval step. To introduce relevant concepts into WM from the KB, we pull the k-hop neighbors of highly salient concepts in WM from the KB (in practice, we find that a hop of 1, where concepts linked directly by a predicate to those in WM, is a sufficient retrieval strategy). Finally, a set of pre-compiled inference rules (section \ref{inference_rules} is applied to WM. Each solution found by applying this rules is used to instantiate a new CG by replacing the postcondition's variables with the values in the solution assignment, and those new CG's are added to WM as newly inferred knowledge.

Because WM is continually updated to include concepts mentioned in the conversation, as well as concepts semantically related to those mentions via predicate links or inferences, our dialogue state is unlikely to diverge from the conversation topic. This strategy is highly flexible, since relevant concepts will populate WM based on fine-grained semantics of each user input, even if the user abruptly changes the subject of discussion.

To avoid a linearly expanding state representation, we implement a threshold-based pruning strategy to limit WM size each turn. Ideally, the most-relevant $k$ concepts in WM would persist to the next turn, whereas less relevant concepts would get pruned. To model the relevance of the concepts in working memory with the current dialogue context, we maintain a real-valued concept-level feature called \textit{salience} that ranges in value between 0 and 1. Each turn of the conversation, concepts mentioned in utterance spoken by either the system or the user are added to working memory with a salience score of 1. At the end of each turn, we decrease the salience of all nodes by a constant value. Additionally, Concepts that are added to WM via inference or knowledge retrieval are not added with a salience of 1, since they are not directly verbalized and therefore are unlikely be of maximal importance to the current discussion. However, salience will be propagated to these concepts based on the salience of their neighbors. We use the following recursive salience formula to compute the salience of concept $i$ with max-salience neighbor $j$:

\begin{equation}
    salience(i) = max(\{salience_{previous}(i), salience(j) - \delta\})
\end{equation}

where $\delta$ is a decay parameter. Using this formula, all concepts with saliences below the $k^{th}$ highest salience value are removed from WM at the end of each turn. In practice, we set $k$ around 100 to minimize latency of expensive graph matching operations on WM. We observe that a relatively low value of $k$ such as this still allows sufficiently large WM to adequately model complex dialogue interactions over many turns.

\subsection{Modeling Truth and Negation}

Another key challenge in human-computer dialogue is the maintenance of truth values over the knowledge present in the dialogue history. Since the utterances shared back-and-forth between two conversational partners can explicitly encode the truth value of the idea being expressed, it is critical that this truth feature is captured and operationalized in a manner that is usable by a dialogue agent. We represent this idea of truth as a positive or negative truth class where each predicate belongs to exactly one of the two truth classes. For instance, the utterance "John is not happy" would produce a predicate \textit{h/happy(John)} that belongs to the negative truth class. If the predicate \textit{h} was instead assigned to the positive truth class, it would correspond to the meaning "John is happy".

Since these truth classes are a feature of the semantic representation employed in our socialbot, they have a direct impact on the dialogue reasoning accomplished through the application of inference rules. The solutions to the inference rule preconditions must match in truth value to that defined within the precondition itself. Namely, preconditions can be specified to only match predicates within a particular truth class. Even if the rest of the predicate structure matches, if the truth class is incompatible, then it is not considered to be a solution to the inference rule precondition. This allows for more fine-grained reasoning over more sophisticated meaning representations of the current dialogue context. 

These truth classes also enable identification of contradictions and corrections. If there are conflicting truth classes on identical predicate structures, then a contradiction can been identified, which is the result of either a misunderstanding or a correction from one of the conversational partners. Although this truth system allows for their identification, strategies for handling and ultimately resolving such situations within dialogue is not addressed in this work and remains as an area for further investigation.

\subsection{Response Selection}

Given an updated WM at the conclusion of all previous understanding, inference, and resolution processes, the socialbot is tasked with identifying the best response to the user's previous utterance. The goal of response selection is to not only choose a relevant response, but also one that is compelling to the user. We define a compelling response as one that signals to the user that the socialbot understood the information shared by the user through their previous utterance while also driving the conversation forward in a meaningful and coherent manner. To this end, we enforce a compound response selection process, where each system response is composed of 2 segments: (1) a reaction to the user's previous utterance and (2) a presentation of some novel information. 

The selection process is the same regardless of which response segment is being selected. A set of candidates are identified from WM and are ranked according to the ranking function. The top candidate is then selected and stored as the value for the appropriate segment. 

\paragraph{Candidate Identification} Each candidate is a set $D$ of predicate instances from WM which represent some distinct thought or piece of information. Content developers of the socialbot are responsible for constructing preconditions (Section \ref{inference_rules}) that capture all valid $D$ that are desired to be handled by the socialbot. Each response precondition is assigned one of two response types \textit{reaction} or \textit{presentation}, following the compound response formulation described previously. We use the graph matching from Section \ref{graph_matching} to identify all candidates of the appropriate type that are present in WM, when response selection is performed for each turn of the conversation.

\paragraph{Ranking Function} Each candidate set $D$ of predicate instances is assigned a score equal to the weighted average of its priority rating and the average salience of the concepts $\in D$, where the priority rating is given a weight of 0.75 and the salience is given a weight of 0.25. The ranking weights and priority ratings were tuned based on manual observation of interactions. The priority rating is determined by the priority class of the precondition that generated $D$. Each precondition is assigned to one of four priority classes: \textit{low, mid, high, critical}. The \textit{low} priority captures responses that have little to no understanding of the current conversational context, and are given a priority rating of 0.1. The \textit{mid} priority captures responses that are generated from at least some partial understanding of the conversational context, and are given a priority rating of 0.4. The \textit{high} priority captures responses that arise from substantial understanding of the conversational context, and are given a priority rating of 0.7. The \textit{critical} priority captures responses that are required in specific situations, such as those dealing with highly sensitive content, and are given a priority rating of 1.0. It is the responsibility of the developer to assign their response preconditions to the most fitting priority class.

%% file: sections/nlg.tex
\section{Natural Language Generation}

Once a collection of predicates have been selected as the response, the final task is to translate the predicates into a natural language English sentence that can be spoken to the user. Our natural language generation module is composed of pairs of preconditions and language templates, where each precondition specifies predicate structures that correspond to the natural language utterance defined in the paired language template.  

We developed a template realization system that allows for dynamic language templates whose tokens can be filled at run-time rather than being pre-specified. Namely, we can generate templates which contain variable slots that are filled on-demand based on variable assignments from the precondition solutions. To ensure that the grammatical constraints of English are met for all produced utterances from this dynamic system, we implemented a grammar-tagging system that allows developers to map grammatical features like tense and plurality of each token to the variables that determine them in the template, if there are such dependencies. At runtime, these grammatical features are then automatically determined and applied based on the variable assignments being applied. We use the python package simplenlg\footnote{\url{https://pypi.org/project/simplenlg/}} which is a port of the SimpleNLG Java API\footnote{\url{https://github.com/simplenlg/simplenlg}} to handle these automatic grammar modifications on English words. 

Both the predicates selected as the reaction and the predicates selected as the presentation are independently processed by this template realization system to produce two English language transcriptions. The final response given by our socialbot to the user is then the concatenation of the reaction transcription with the presentation transcription.

%% file: sections/interaction_design.tex
\section{Dialogue Interaction Design}

Designing our chatbot's interactions and content involves creating a number of handcrafted knowledge elements. Specifically, we developed a set of CG's defined using structured text files and a custom CG compiler to fill out our chatbot's prior knowledge base, inference rules, and NLG templates. Although handcrafted elements are costly to develop, we find they are the most reliable and efficient way to build a satisfying chatbot while testing novel dialogue management strategies. The remainder of this section summarizes the internal guidelines we used to develop the handcrafted elements of our chatbot. These guidelines should not be taken as an empirically verified conversation theory, since they are based on our experience as chatbot designers and not on quantitative analyses.

\subsection{Initiative Design}

Table \ref{tab:interactions} outlines a coarse-grained categorization scheme we used to facilitate proper initiative modelling in our bot's interactions. The categorization roughly divides types of interactions by which speaker has more control over the conversation subject. We observe from experience in this and prior competitions that chatbots often dominate conversations, with bot utterances accounting for the vast majority of all tokens spoken in the conversation. This leads to a conversation style that feels overly scripted and interrogative. To mitigate this, we encouraged development of inference rules and response templates that produce interactions that are evenly distributed among our four categories. 

\begin{table}[]
    \setlength\extrarowheight{2pt}
    \centering
    \begin{tabular}{l||l|l}
         & Bot Initiative & User Initiative\\
         \hline\hline
         \multirow{10}{*}{Strong} & \textbf{Prompts} & \textbf{Answer the User}\\
         & Objectives: & Objectives: \\
         & \hspace{5mm} Elicit meaningful information from user. & \hspace{5mm} Address the received prompt. \\
         & \hspace{5mm} Build rapport with the user. & \hspace{5mm} Provide compelling information \\
         & \hspace{5mm} Identify user's interests. & \hspace{5mm} to enable user follow-ups. \\
         & Example: & Example: \\
         & \hspace{5mm} What did you like about the movie? &  \hspace{5mm} User: What’s your favorite movie? \\
         & & \hspace{5mm} Bot: I like avengers because I \\
         & & \hspace{5mm} relate to Wanda's feelings \\
         & & \hspace{5mm} of not fitting in. \\
         \hline
         \multirow{10}{*}{Weak} & \textbf{Share Experiences} & \textbf{Reaction} \\
         & Objectives: & Objectives: \\
         & \hspace{5mm} Peak the user's interest in the bot & \hspace{5mm} Ensure the user feels understood. \\
         & \hspace{5mm} as a relatable entity. & \hspace{5mm} Take initiative back if user is unsure \\
         & \hspace{5mm} Signal to the user that they & \hspace{5mm} how to proceed or a compelling \\
         & \hspace{5mm} can take  initiative. & \hspace{5mm} conversation direction is identified. \\
         & Example: & Example: \\
         & \hspace{5mm} Bot: I always thought sashimi would be gross & \hspace{5mm} User: I bought my first house! \\
         & \hspace{5mm} but I just tried it and it was delicious! & \hspace{5mm} Bot: Congratulations! \\
         & & \hspace{5mm} That sounds so exciting. \\
    \end{tabular}
    \vspace{3mm}
    \caption{Outline of conversation interaction strategies employed in the bot.}
    \label{tab:interactions}
\end{table}

\paragraph{Strong System Initiative Interactions} are prompts or questions, where the bot directs the conversation and is likely to maintain the initiative even after the user has responded. This is the easiest type of dialogue interaction to construct since the bot has the strongest degree of control over the conversation. The goal is to elicit information from the user about their opinions and preferences, which can then be used to further personalize the conversation to the user's interests. For example, the bot may prompt the user "Do you watch movies a lot". Based on whether the user answers yes or no, the bot can determine whether to follow up with more movie-related interactions or to move onto another topic. This interaction style is the most commonly seen among existing chatbots, since it allows the chatbot to constrain the expected response space for easier dialogue management.

\paragraph{Weak System Initiative Interactions} occur when the bot has conversational control but does not necessarily maintain the initiative after its interaction is given. In these cases, the bot generally shares information about its own opinions and experiences without giving a direct prompt, allowing the user the freedom to follow up with a response that may or may not take the initiative for themselves. For example, if the bot shares "Avengers is my favorite movie of all time", the user could take the initiative by prompting the bot for more information with a question or even by switching to a different topic of conversation. On the other hand, it is also natural for the user to decline taking the initiative via a short acknowledgment of the bot's experience (e.g. "That's cool"). We find that users are more engaged when these weak initiative presentations are mixed with strong initiative prompts, since it breaks up an interrogative conversation style and gives the user a larger degree of freedom in how they respond. 

\paragraph{Strong User Initiative Interactions} are symmetrical to strong system initiative interactions in that the user prompts the bot for some information and is expecting to maintain control of the conversation even after the bot has given its response. In this situation, the bot's only responsibility is to respond to the prompt. These responses provide a compelling opportunity for the bot to engage the user with its unique experiences and personality. Additionally, these responses should also provide enough information so that the user can follow up with additional questions or reactions. For example, if the user asks "Do you have a pet", the bot should fully answer the question and include additional information the user might find interesting (e.g. "I have a dog named Nyla. She is three years old and we love playing fetch together").

\paragraph{Weak User Initiative Interactions} are characterized by the user sharing information without a direct prompt to the bot. Although in these situations it would likely be natural for the bot to either accept or decline the initiative, we chose for our bot to claim the initiative in the majority of cases. For example, if the user shares an experience such as "I just bought my first car", it would be natural for the bot to just react with a congratulatory expression (e.g. "Oh, congratulations. That's exciting!") and leave the initiative with the user. However, we generally designed our bot to transition to strong system initiative by following this reaction with a prompt, such as "What is your favorite thing about the car". This was done in order to ensure the conversation always felt like it was moving forward from the user's perspective. In the future, we plan to design our bot to take the initiative less frequently by improving the natural language understanding and topic coverage in order to better handle longer stretches of user initiative interactions.

Taken together, this initiative-based dialogue interaction design features frequent "weak-initiative" responses that allow the user to take initiative if they have follow up discussion points or additional questions for our bot. However, our bot also has a continuous supply of prompts and discussion points generated from its inference system to carry the conversation forward if the user does not wish to take these opportunities.

\subsection{Topic Design}

Although there is no discrete notion of topics in our system, we made an effort to focus on a number of popular conversations that we believe users are likely to be engaged with. These topics are realized as simply a collection of concepts, predicates and implication rules in the  KB. As described in section \ref{working_memory}, dialogue management is handled on a fine-grained level allowing for fluid, non-discrete transitions between the content in these topics.

\input{topics/animal}
\input{topics/art}
\input{topics/college}
\input{topics/cuisine}
\input{topics/employment}
\input{topics/movies}
\input{topics/postpandemic}
\input{topics/sports}
\input{topics/technology}
\input{topics/travel}
\input{topics/videogame}

%% file: topics/animal.tex
\subsubsection{Animals}

The animals component covers both a discussion on the user's pets as well as general animals interactions. The pets discussion focuses on learning more about the temperament and activities enjoyed by the user's pets. The animals discussion elicits the user's favorite animal and their experiences going to the zoo. Our bot also shares her own experiences with her pet, a dog named Nyla, and her favorite exhibit to see at the zoo.

%% file: topics/art.tex
\subsubsection{Art}

The art component covers interactions about artists and art-related activities. Our bot cares about and presents her curiosity towards users' opinions and experience centering around art. Topics involved include going to exhibitions, digital drawing, and motivation to be an artist if users consider themselves as artists. Our bot also brings up her own perspectives about art in her interactions. For example, her favorite artist is Claude Monet because she appreciates his use of colors in the paintings. This component is meant to not only engage users in entertaining conversations but also express Our bot's personalities.

%% file: topics/college.tex
\subsubsection{College}

The college component focuses on conversations with past, current, and prospective college students. For past college students, interactions center around differences between colleges in the past and colleges nowadays. For current college students, our bot interact with users by discussing opinions and experience on feelings of remote learning as well as going back to the campus. For prospective college students, our bot is mainly curious about users' expectations for college and progress of college application. This component is also designed to cover more common college-related topics, such as majors, experience in different school years, and college life. With this component, our bot resonates with users who share similar college experience and embraces fresh minds as well.

%% file: topics/cuisine.tex
\subsubsection{Cuisine}

The cuisine component allows our bot to handle interactions about cuisines by geographical area and food type as well as cuisine-related activities. Three major cuisines by geographical areas include Asian, European, and American cuisines. Furthermore, our bot is capable of discussing different cuisines by country under each of the three major cuisines. Interactions about various types of food, such as drinks, fast food, fruits, and vegetables are also covered. These interactions are achieved by building an extensively comprehensive ontology of the consumable. One feature that our bot is capable of doing is asking for descriptions after identifying unknown food or drinks. Our bot also interacts with users on topics of cuisine-related activities, including cooking, baking, having meals, and going to restaurants. As one of the most common daily activities, this component involves topics that everyone can relate to and talk about.

%% file: topics/employment.tex
\subsubsection{Employment}

The employment component is designed to cover topics centering around jobs and work-related activities. Our bot handles different types of jobs when brought up by users with her own opinions. For example, she respects people who work in the technology sector and recalls her memory about dreaming of being a jewelry designer when users claims themselves to work as designers. When users mention jobs that our bot does not know, she asks for explanations. Our bot is also curious about users' attitudes towards their jobs. Other interactions cover work-related activities, such as going back to the office and working in teams. This component relates users who have work experience as one of the most common topics people care about.

%% file: topics/movies.tex
\subsubsection{Movies}

The movies component has content across 4 broad areas: movies, television shows, streaming providers, and miscellaneous related content. The movies content is centered around Marvel and Disney movies, but includes extensive handling for unknown movies. The television content focuses on different T.V. genres. The streaming providers content deals with a variety of available streaming platforms. The miscellaneous content is focused on both relating to users and attempting to bring up questions in other components to further the dialogue.

%% file: topics/postpandemic.tex
\subsubsection{Post-Pandemic}

The Post-Pandemic component is meant to interact with the user on how life will have changed after the pandemic. Part of it is meant to elicit a question form the user along the lines of "How has the pandemic changed your life", to which our bot shares her experience. The component also addresses and interacts with user about the shift to remote work, as well as a greater appreciation for human interaction post-lockdown. This component is meant to relate to users in what has been a shared experience  for most people.

%% file: topics/sports.tex
\subsubsection{Sports}

The sports component covers several common interactions. The content is meant to get the user talking about not only their opinions but also their emotions and experiences related to sports. Our bot also shares her sports experiences and feelings, with focus on teammates, friends, and family. Our bot can handle common topics such as sports teams, wanting to play / having played sports, and watching sports events.

%% file: topics/technology.tex
\subsubsection{Technology}

The technology component is meant to cover 3 main areas: Computers, Phones, and Artificial Intelligence. The Computer section focuses on choices like PCs vs Macs, Linux/MacOS/Windows, and other computer enthusiast content, like building your own computer. Here the goal is to interact with PC enthusiasts and share our bot's own experience with technology. The phone section talks about the differences between IOS and Android, as well as other user preferences, like style over performance. Finally, the AI component interacts with the user about opinions on AI, its dangers and utility. The goal is to provide a "meta" interaction, where an AI chatbot talks with the user about AI. The technology component also has a range of interactions for almost any electronic device, providing broad coverage for tech discussions.

%% file: topics/travel.tex
\subsubsection{Travel}

The travel component focuses on traveling experience about trips and destinations. Interactions about trips include opinions on past, current, or future trips and things users like to do for trips. With an ontology that covers common cities and attractions in the United States and major cities from other countries, our bot is able to discuss perspectives and experience with users about traveling to different destinations. Our bot also shares her own stories and expresses her personalities when interacting with users. For example, she loves going to beaches and prefers to live in small cities when getting old. This component is designed to encourage engaging conversations about traveling.

%% file: topics/videogame.tex
\subsubsection{Videogames}

The video games component focuses on conversation with users interested in video games. Our bot aims to create back and forth exchanges by asking the user about their favorite games and sharing her own preferences. Our bot also covers different aspects of the gaming experience learning more about the user’s favorite game genres and gaming devices. The component aims to provide any video game player with a place to converse about their hobby.

%% file: sections/discussion.tex
\section{Discussion}

Unlike other symbolic approaches, the main advantage of our approach is the flexibility it provides in transitioning between different discussion points. There is no notion of a scripted dialogue pathway in our approach, since each interaction is triggered based on the satisfaction of a set of logical conditions encoding its appropriateness to the current conversation context. This means that any interaction can be triggered from a combinatorial number of conversation contexts, and there is no strict order in which interactions are presented to the user. 

\begin{figure}[htp]
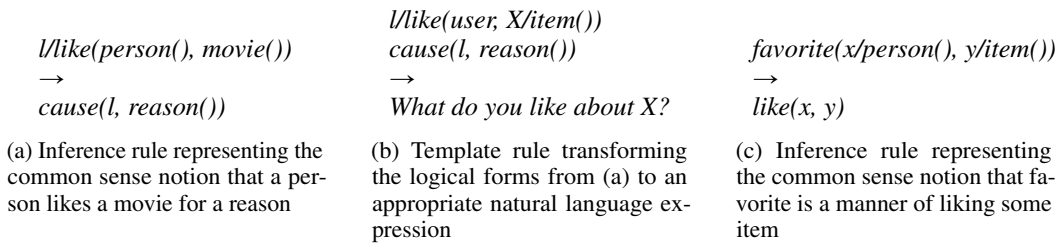

\centering
\subcaptionbox{Inference rule representing the common sense notion that a person likes a movie for a reason}[0.3\textwidth]{
\begin{tabular}[t]{l} 
    \textit{l/like(person(), movie())} \\
    $\rightarrow$ \\
    \textit{cause(l, reason())}
\end{tabular}
}
\hfill
\subcaptionbox{Template rule transforming the logical forms from (a) to an appropriate natural language expression}[0.3\textwidth]{
\begin{tabular}[t]{l} 
    \textit{l/like(user, X/item())} \\
    \textit{cause(l, reason())} \\
    $\rightarrow$ \\
    \textit{What do you like about X?}
\end{tabular}
}
\hfill
\subcaptionbox{Inference rule representing the common sense notion that favorite is a manner of liking some item}[0.3\textwidth]{
\begin{tabular}[t]{l} 
    \textit{favorite(x/person(), y/item())} \\
    $\rightarrow$ \\
    \textit{like(x, y)}
\end{tabular}
}
\hfill
\caption{Example inference and template rules}
\label{dialogue_rules}
\end{figure}

\begin{figure}[htp]
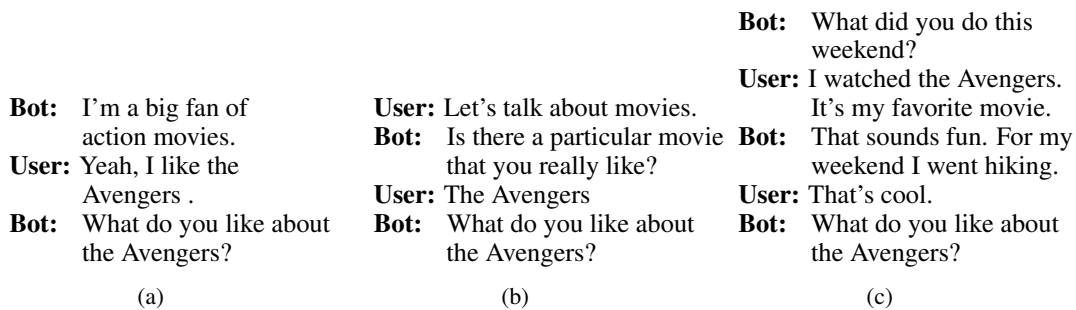

\centering
\subcaptionbox{}[0.3\textwidth]{
\begin{tabular}[t]{l} 
    \textbf{Bot:} \hspace{3pt} I'm a big fan of \\
    \hspace{25pt} action movies. \\
    \textbf{User:} Yeah, I like the \\
    \hspace{25pt} Avengers . \\
    \textbf{Bot:} \hspace{3pt} What do you like about \\
    \hspace{25pt} the Avengers? \\
\end{tabular}
}
\hfill
\subcaptionbox{}[0.3\textwidth]{
\begin{tabular}[t]{l} 
    \textbf{User:} Let's talk about movies. \\
    \textbf{Bot:} \hspace{3pt} Is there a particular movie \\
    \hspace{25pt} that you really like? \\
    \textbf{User:} The Avengers \\
    \textbf{Bot:} \hspace{3pt} What do you like about \\
    \hspace{25pt} the Avengers? \\
\end{tabular}
}
\hfill
\subcaptionbox{}[0.3\textwidth]{
\begin{tabular}[t]{l} 
    \textbf{Bot:} \hspace{3pt} What did you do this \\
    \hspace{25pt} weekend? \\
    \textbf{User:} I watched the Avengers. \\
    \hspace{25pt} It's my favorite movie. \\
    \textbf{Bot:} \hspace{3pt} That sounds fun. For my \\
    \hspace{25pt} weekend I went hiking. \\
    \textbf{User:} That's cool. \\
    \textbf{Bot:} \hspace{3pt} What do you like about \\
    \hspace{25pt} the Avengers? \\
\end{tabular}
}
\hfill
\caption{Example conversations}
\label{dialogue_paths}
\end{figure}

For example, one of our chatbot's interactions is to ask the user why they enjoyed a particular movie, encoded in the inference rules and template rules shown in Figure \ref{dialogue_rules}. As long as the bot knows that the user likes the movie, this is an appropriate interaction that can take place, regardless of the specific conversation context that led to this knowledge. Figure \ref{dialogue_paths} shows 3 example conversation flows in which this interaction is applicable and would be triggered by the context. 

Conversation (a) is straightforward: the user volunteers that they like a particular movie, thus directly triggering the rule and its corresponding NLG template (b), resulting in the generation of candidate response "What do you like about the Avengers". On the other hand, conversations (b) and (c) demonstrate that it is also possible to trigger this interaction by aggregating information across multiple turns using reasoning. 

In conversation (b), the knowledge that the user likes Avengers is the product of the user's response to a question posed by the chatbot. Once this question is resolved to the user's answer (see Section \ref{references}), the resulting predicate representation is equivalent to that constructed from the user utterance of "I like the Avengers" from conversation (a). This results in the satisfaction of rule (a) and, as a consequence, the question "What do you like about the Avengers" will be generated as a response candidate.

Conversation (c) illustrates yet another reasoning process that produces the question as a response. In this example, rule (c) enables inferring that the user likes Avengers from the user sharing that it is their favorite movie. Through this common sense reasoning process, rule (a) can once again be satisfied, resulting in the generation of the same candidate response from template rule (b). Notice that in this last example the response "What do you like about the Avengers" is not immediately chosen as the follow-up. This is often the case, since any particular turn of dialogue can result in many response candidates. In conversation (c), the response "That sounds fun. For my weekend..." was chosen by the response selection ahead of the question, but this does not necessarily invalidate the appropriateness of asking the question in the future. Since the inference rules driving the question are still satisfied, the question is still recognized as appropriate, leading to the behavior seen in conversation (c) where the bot circles back to further discuss the user's favorite movie. 

As demonstrated by the previous examples, although our dialogue management approach is symbolic or "rule-based", it's ability to reason over fine-grained predicates to produce candidate responses yields a massive amount of flexibility in the conversations the the bot is capable of, while maintaining the explainability and controllability of a symbolic state representation. Since each interaction is triggered based on a set of logical conditions checking its appropriateness, conversation design from a developer perspective is less about constructing scripts and more about creating interesting interactions that capitalize on the information shared by each conversational partner.

\subsection{Implementation Challenges}

The effort described in this work details an initial step into the development of an inference-driven dialogue management approach. Although this approach affords many benefits as discussed previously, there are a handful of ongoing challenges to its implementation that we outline here. 

First, it is crucial to this approach to converge on a semantic representation for the Concept Graph's predicate structures that is suitable for the intended conversation domain. Our semantic representation was inspired by AMR \citep{amr} with additional added features, such as tense. Although use of a robust AMR parser would have been ideal for our NLU step, performance of current AMR parsers is not robust enough, particularly when generalizing to new datasets, to be a reliable solution for our conversation domains. Therefore, we decided to translate more reliable dependency parse outputs into the semantic representation we desired. This incurred a substantial cost in terms of development effort as well as ongoing compatibility difficulties when we modified our semantic representation, since modifications invalidated large portions of content (e.g. inference rules and templates) that had already been created.

A second major challenge is that our approach heavily relies on the efficient satisfaction checking of hundreds or thousands of inference rules against the Working Memory Concept Graph. This is a significant computational challenge, since the graph-matching problem required parallel checking of each rule in order to have a reasonable performance. To our knowledge, there are no existing GPU implementations of graph matching for checking a large number of query graphs in parallel, so we had to incur the cost of constructing our own in-house GPU-based graph-matcher.

Lastly, although content development is relatively efficient for a well-trained developer, the content creation process has a steep learning curve and is not intuitive to newcomers. Although content design using the inference rules and templates is highly powerful for constructing intelligent and compelling dialogue behaviors, it is fairly challenging for novice developers to understand how to get the socialbot to do what they want. The idea of constructing interactions that are triggered by the bot's built-in knowledge and reasoning process is simply less intuitive than the script-style conversation design of state machine-based dialogue frameworks.

\subsection{Future Work}

Our future efforts will focus on two main directions. First, although this current iteration involved many handcrafted elements, there is no necessary relationship between inference-driven dialogue as an approach and handcrafted design elements. Our future efforts will focus on replacing the handcrafted elements with scalable models that accomplish similar tasks in a more robust and less developer-involved manner. Specifically, we look to train a neural parser for our semantic representation rather than utilize transformation rules, to incorporate a neural inference model that produce the inferences that drive dialogue, to select responses based on a reinforcement-learning paradigm, and to replace the template-based NLG module with a neural graph-to-text approach for translating response predicates into natural language. As a second direction for future work, we aim to use our approach as a platform to investigate linguistic theories of dialogue. The high degree of controllability and explainability of the symbolic elements of our approach coupled with the flexibility of our dialogue management framework should provide a powerful computational lens for this area.

%% file: sections/conclusion.tex
\section{Conclusion}

In this report we presented an approach and implementation to chat-oriented human computer dialogue. Our implemented chatbot is a realization of modelling conversation as a collaborative inference process between two speakers. It uses a symbolic concept representation and inference engine to generate novel ideas on the fly each dialogue turn, thereby producing interesting responses that logically cohere with the conversation context.